# Fast Belief Update Using
# Order-of-Magnitude Probabilities

**Moisés Goldszmidt**
Rockwell Science Center
444 High Street
Palo Alto, CA 94301
moises@rpal.rockwell.com

## Abstract

We present an algorithm, called *Predict*, for updating beliefs in causal networks quantified with order-of-magnitude probabilities. The algorithm takes advantage of both the structure and the quantification of the network and presents a polynomial asymptotic complexity. *Predict* exhibits a conservative behavior in that it is always sound but not always complete. We provide sufficient conditions for completeness and present algorithms for testing these conditions and for computing a complete set of plausible values. We propose *Predict* as an efficient method to estimate probabilistic values and illustrate its use in conjunction with two known algorithms for probabilistic inference. Finally, we describe an application of *Predict* to plan evaluation, present experimental results, and discuss issues regarding its use with conditional logics of belief, and in the characterization of irrelevance.

## 1 Introduction

Order-of-magnitude probabilities (OMPs) and their formal relatives, $\epsilon$-semantics and kappa calculus, have had considerable impact in knowledge representation in the specific areas on nonmonotonic reasoning, belief revision, and the representation of uncertainty [12, 15, 9, 18, 25]. The reasons for this success are various: They allow the representation of belief and uncertain knowledge as a set of if-then rules, they provide a well known mechanism for belief update which is linked to notions of conditioning, and finally the provide a link to probability theory [16], conditional logics [3], and other calculi for uncertainty such as possibility theory [1]. Researchers in uncertainty in AI had hoped that this abstraction of numerical probabilities would also open the doors for faster algorithms for belief update, in addition to better knowledge representation methods. Although Goldszmidt and Pearl [16] introduced a semi-tractable algorithm for OMPs, the algorithm did not run on distributions represented by belief networks, a favorite representation among Bayesian practitioners.

In this paper we introduce a polynomial algorithm for prediction tasks in networks quantified with order-of-magnitude approximation probabilities. The algorithm, which we call *Predict* has essentially the same complexity as the polytree algorithm [23], with the additional advantage that *Predict* is not bothered by undirected cycles in the network. *Predict* exhibits a conservative behavior in the sense that it is sound, but not complete. Soundness means that when the algorithm yields a *believed value* for a variable, a manipulation of the OMP according to their properties will yield the same result. Incompleteness means that sometimes the algorithm will fail to recognize a believed value for a particular variable. In this paper we characterize instances where *Predict* is complete, present a polynomial algorithm for testing sufficient conditions for completeness, and introduce a stratified procedure for computing a complete set of believed values for each variable.

In addition we study the role of *Predict* as a polynomial estimator of probabilistic values, for deciding whether $P(x) > \epsilon$ for a given real valued $\epsilon$, and illustrate the use of *Predict* for speeding up two known algorithms for probabilistic belief update: bounded-conditioning proposed by Horvitz, Suermondt, and Cooper [19], and the search-based algorithm proposed by Poole [24]. We also present experimental results and describe an application to plan evaluation.

This paper is organized as follows. Section 2 reviews the main concepts behind OMPs and network representations. Section 3 introduces *Predict* and its main properties, including theorems about soundness and completeness, and a stratified procedure for completeness. Section 4 discusses *Predict* in the role of performing approximate probabilistic inference, shows its use in probabilistic update algorithms, and describes an application to plan evaluation including experimental results. Section 5 discusses *Predict* in the context of providing an algorithm for inference and representation of conditional logics of belief in networks, a new notion of irrelevance, and finally summarizes the main results and future challenges.

## 2  OMPs, Rankings, and Networks of Belief

To see the relation between $\kappa$-ranking functions [26], $\epsilon$-semantics [12, 23], and probabilities, imagine an ordinary probability function P defined over a set $\Omega$ of possible worlds (or states of the world), and let the probability $P(\omega)$ assigned to each world $\omega$ be a polynomial function of some small positive parameter $\epsilon$, for example, $\alpha$, $\beta\epsilon$, $\gamma\epsilon^2$, ..., and so on. Accordingly, the probabilities assigned to any proposition $x$, as well as all conditional probabilities $P(x|y)$, will be rational functions of $\epsilon$. Now define the function $\kappa(x|y)$ as the lowest $n$ such that $\lim_{\epsilon \to 0} P(x|y)/\epsilon^n$ is nonzero. In other words, $\kappa(\mathbf{x}|\mathbf{y}) = n$ is of the same order of magnitude as $P(\mathbf{x}|\mathbf{y})$.

It is easy to verify that when $\epsilon$ is infinitesimally close to zero, the following properties of $\kappa$ can be derived from the analogous properties of P:

1. $\kappa(\mathbf{x}) = \min_i\{\kappa(\omega_i)|\omega_i \models \mathbf{x}\}$
2. $\kappa(\mathbf{x}) = 0$ or $\kappa(\neg \mathbf{x}) = 0$, or both
3. $\kappa(\mathbf{x} \vee \mathbf{y}) = \min\{\kappa(\mathbf{x}), \kappa(\mathbf{y})\}$
4. $\kappa(\mathbf{x} \wedge \mathbf{y}) = \kappa(\mathbf{x}|\mathbf{y}) + \kappa(\mathbf{y})$

Note that these properties reflect the usual properties of probabilistic combinations (on a logarithmic scale) with *min* replacing addition, and addition replacing multiplication.

If we think of $n$ for which $P(\omega) = \epsilon^n$ as measuring the degree to which the world $\omega$ is disbelieved, then $\kappa(\mathbf{x}|\mathbf{y})$ can be thought of as the degree of disbelief (or surprise) in $\mathbf{x}$ given that $\mathbf{y}$ is true. In particular, $\kappa(\mathbf{x}) = 0$ means that $\mathbf{x}$ is a serious possibility or *plausible* and $\kappa(\neg\mathbf{x}) > 0$ translates to "$\mathbf{x}$ is believed."[1] Note that in this case, $P(\neg\mathbf{x})$ will approach zero in the limit as $\epsilon$ approaches zero and consequently $P(\mathbf{x})$ will approach one. The algorithm we propose in this paper allows for an efficient computation of the plausible values of a variable $x$.

### 2.1  Network Representations

One of the most efficient ways to representing and reasoning with a probability distribution is through the use of Bayesian networks. A network consist of a directed acyclic graph $\Gamma$ and a quantification $\mathcal{Q}$. The nodes in $\Gamma$ represent the variables of interest in the domain, and the edges represent direct (causal) influences. In this paper we will use lowercase letters of the alphabet, such as $x$, to represent both nodes and variables, and $\mathbf{x}$ will denote an instantiation of $x$ (i.e., the value of $x = \mathbf{x}$).[2] If $x$ is a node in $\Gamma$, then $\pi(x)$ will denote the set of parents of $x$ in $\Gamma$. $\vec{\pi}(x)$ denote an instantiation of the parents of $x$, and when use inside $P(x|\pi(x))$ or $\kappa(x|\pi(x))$, $\pi(x)$ denotes the conjunction of the parents of $x$.[3] The quantification $\mathcal{Q}$ of $\Gamma$ specifies the strength of the influence of the parents of $x$, $\pi(x)$, on $x$. In probabilistic reasoning, this quantification is done in terms of conditional probabilities $P(x|\pi(x))$. Furthermore, the structure of the network encodes a set of conditional independence assumptions that translate into the following equation:

$$P(x_1,\ldots,x_n) = \prod_{1 \leq i \leq n} P(x_i|\pi(x_i)) \qquad (1)$$

where $x-1,\ldots,x_n$ are the nodes in $\Gamma$. The polytree algorithm [23] takes advantage of these independencies to update a probability distribution in polynomial time when the underlying graph is a polytree, where $\Gamma$ is a polytree iff $\Gamma$ does not contain undirected cycles. If $\Gamma$ contains undirected cycles the problem of updating the probability distribution in a Bayesian network is NP-hard [4] and all (known) exact algorithms are exponential.

A network can be also quantified with $\kappa$-rankings by providing local conditional rankings $\kappa(x|\pi(x))$, representing the degree of disbelief (or an order of magnitude probability) on the state of $x$ given the values taken by its parents $\pi(x)$ in the network $\Gamma$ [15, 20]. We can then draw a parallel to Eq. 1 in terms of kappa rankings $\kappa(x_1,\ldots,x_n) = \sum_{1 \leq i \leq n} \kappa(x_i|\pi(x_i))$. Furthermore, it was shown by Hunter [20] that the polytree algorithm is also valid in a network quantified with rankings. Thus, updating the ranking of a variable $x$ can be done quite efficiently if $\Gamma$ is a polytree, yet the computation becomes exponential once more, if $\Gamma$ contains undirected cycles. However, as the next section will show, we can compute the plausible values of each node $x_i$, that is those values $\mathbf{x_i}$ such that $\kappa(\mathbf{x_i}) = 0$, quite efficiently.

## 3  Predicting Belief Change

The algorithm, called *Predict*, is shown in Figure 1. It takes as input a network of belief, and computes the plausible values for each node $x_i$. We call this set of values the *plausible set* of $x_i$. Both plausibility and the plausible set of a variable are defined next.

**Definition 1 (Plausibility and Belief.)** *Given a ranking $\kappa$, we say that $\mathbf{x}$ is plausible given $\mathbf{y}$, denoted $Pl(\mathbf{x}|\mathbf{y})$, iff $\kappa(\mathbf{x}|\mathbf{y}) = 0$. Similarly, we say that $\mathbf{x}$ is believed given $\mathbf{y}$, denoted $Bel(\mathbf{x}|\mathbf{y})$, iff $\kappa(\mathbf{x'}|\mathbf{y}) > 0$*

---

[1] We are assuming that the variable $x$ is propositional. The generalization to multiple valued variables is straightforward.

[2] Variables don't have to be propositional, yet, if $x$ is a propositional variable $\neg\mathbf{x}$ denotes the negation of $\mathbf{x}$.

[3] We will keep the usual notation in probabilities $P(x,y)$ ($\kappa(x,y)$) to denote $P(x \wedge y)$ ($\kappa(x \wedge y)$).

PROCEDURE *Predict*
Input: An OMP network $\Gamma$.
Output: PlSet($x_i$) = $\{x_j | Pl(x_j)\}$ for each node $x_i$ in $\Gamma$.

1. Let $Q = (x_1, \ldots, x_n)$ be a topologically sorted list of the nodes in $\Gamma$.

2. While $Q$ is not-empty

   2.1 Remove the first node $x_i$ from $Q$; If $x_i$ is a root node then PlSet($x_i$) = $\{x_j | \kappa(x_j) = 0\}$

   2.2 Else let $\{x_r, \ldots, x_s\}$ be the set of $x_i$'s parents; then PlSet($x_i$) = $\{x_j\}$ for all values $x_j$ of $x_i$ such that

   $$\kappa(x_j | \vec{\pi}(x_i)) + \sum_{r \leq p \leq s} \kappa(x_p) = 0 \qquad (2)$$

END *Predict*

Figure 1: *Algorithm for computing the plausible values of each node $x_i$ in an OMP network.*

for every value $x'$ different than $x$.[4] The plausible set *of $x$, written* PlSet($x_i$) *is defined as the set of values $\{x\}$ such that* Pl(x).

The algorithm traverses the nodes in the network following any topological sort of these nodes, and computes PlSet($x_i$) based on the plausibility of $x_i$'s parents (see Eq. 2). Thus, the asymptotic complexity of the algorithm is determined by the number of nodes in the network and the number of edges (Step 2.2). Let $E$ denote the number of edges in the network, $N$ the number of nodes, and let $LU$ represent a look-up operation local to each family in the network.[5] It is easy to verify that the complexity of *Predict* is given by the following theorem:

**Theorem 1** *The complexity of Predict is $O(E + N \times LU)$.*

Note that this is the same complexity of the polytree algorithm [23], except that *Predict* also applies to networks containing undirected cycles. As described in Figure 1, *Predict* does not admit any evidence. A minimal change would allow the incorporation of evidence on roots, and the representation of actions and decisions for evaluating plans. For the first case we simply take into account the evidence when computing Step 2.1 in Figure 1. For incorporating changes due to actions and decisions, we simply modify the graph and set the new roots to be the direct children (in the network) of the actions or decisions [8].

We use well defined notions of soundness and completeness to characterize the output of *Predict* in terms of $\kappa$-rankings. The interplay with probabilities is characterized in Section 4.

[4] In relation to infinitesimal probabilities, Bel($x|y$) implies that P($x|y$) approaches one as $\epsilon$ approaches zero.

[5] Recall that for each family in the network we must have the conditional ranking $\kappa(x_i | \pi(x_i))$.

**Definition 2 (Soundness)** *Given a ranking $\kappa$ represented by an OMP network. We say that a run of Predict is sound iff the fact that $x$ is not in* PlSet(x) *implies that $\kappa(x) > 0$.*

**Theorem 2** *Predict is sound for any OMP network.*

The notion of completeness establishes the opposite: If $\kappa(x_j) > 0$, then we would like for $x_j \notin$ PlSet($x_i$). In other words, if the ranking represented by the network establishes that a certain value of a variable is believed, then running *Predict* on this network should yield a plausible set for this variable containing a unique element corresponding to this believed value.

**Definition 3 (Completeness.)** *Given a network of belief we say that a run of Predict is complete iff the fact that $\kappa(x) > 0$ implies that $x$ is not in* PlSet(x).

Unfortunately, *Predict* is not complete in general. What incompleteness means is that sometimes *Predict* will regard more than one value of a node as plausible, even when there is enough information in the quantification of the network to determine which of these values are not plausible. The source of the incompleteness is Eq. 2 in Figure 1. This equation approximates the value of $\kappa(x_i)$ by assuming that $x_i$'s parents are *irrelevant* to each other (see Def. 4). There are however, important classes of networks for which *Predict* computes a complete set of plausible values for each node. These cases, described in the next section, depend on both the structure of the network, and its quantification.

### 3.1 Conditions for Completeness

The computations of *Predict* will be *complete* precisely whenever the computations in Eq. 2 will yield the same set of plausible values than equation $\kappa(x_j | \vec{\pi}(x_i)) + \kappa(\vec{\pi}(x_i))$ for each node $x_i$. In this section we provide a set of sufficient conditions for completeness by focusing on conditions for $\kappa(\vec{\pi}(x_i)) = \sum_{r \leq p \leq s} \kappa(x_p) = 0$ (where $\{x_r, \ldots, x_s\}$ is the parent set of $x_i$). We also present a polynomial $O(E)$ algorithm for checking these conditions. A stratified method for refining an initial run of *Predict* until completeness is achieved is described in Section 3.2. Consider the following definition:

**Definition 4 (Irrelevance.)** *Given a ranking $\kappa$, a set of variables $\{x_1, \ldots, x_n\}$ are irrelevant to each other iff whenever $\kappa(x_1) = \cdots = \kappa(x_n) = 0$ then $\kappa(x_1, \ldots, x_n) = 0$ (for all the instantiations of $\{x_1, \ldots, x_n\}$).*

This definition is by all means preliminary,[6] but it serves its purpose for this paper, namely to provide a minimal language for establishing conditions for the completeness of *Predict*, and for describing a stratified algorithm for completeness in Section 3.2.

As Theorem 5 below points out, the assumption of irrelevance in Eq. 2 is true only under certain conditions. We need the definition of a *backpath*.

**Definition 5 (Backpaths)** *Let $x$ and $y$ be two nodes in a network $\Gamma$. Let $BP$ be an undirected path between $x$ and $y$ such that all nodes in $BP$ are ancestors of either $x$ or $y$. We call $BP$ a* **backpath**. *We say that the backpath is* **blocked***, iff there exists at least one node $b$ in $BP$ such that $Bel(b)$ for some value $b$ of $b$.*

Note that the definition of a blocked backpath not only depends on the structural configuration of the network, but also on the particular quantification of the network. As we will see, completeness will also depend on the particular quantification (and sometimes on the evidence) in the network. The reason is that the quantification (and the evidence) will induce belief changes that can in turn produce blocked paths.

**Theorem 3** *Let $\kappa$ be a ranking represented by a network with structure $\Gamma$, and let $x$ and $y$ be two nodes in the network such that all backpaths between $x$ and $y$ are blocked, then $x$ and $y$ are irrelevant in the sense of Def. 4.*

Thus, if the structure of the network does not contain cycles, or if the quantification (or evidence) "block" all backpaths then the computation in Eq. 2 is always valid and *Predict* is complete.

**Theorem 4** *If the network does not contain undirected cycles, then Predict is complete.*

**Theorem 5** *If a run of Predict returns a set of believed nodes blocking all backpaths then the run of Predict is complete.*

These results point to a procedure for checking whether a particular run of *Predict* is complete. This procedure takes as input the set of believed nodes. It removes their outgoing edges and then runs a breadth-first search algorithm to construct a spanning forest. If no cross-edge is detected (i.e., there are no cycles) the run of *Predict* is complete. If a cycle is reported, then the run may be incomplete, since there are unblocked paths. Note however that unblocked paths do not imply incompleteness. The complexity of this procedure is $O(E)$.

Finally, there is a special quantification of the network that will guarantee completeness. This quantification is equivalent to the one studied by Poole [24].

**Definition 6** *Let $\Gamma$ be a network and let $\mathcal{Q}$ be its quantification representing $\kappa$-ranking. We say that $\mathcal{Q}$ is* definite *iff for every $x_i \in \Gamma$ and every instantiation $\bar{\pi}(x_i)$, there exists a unique instantiation $\mathbf{x}_i$ such that $\kappa\big(\mathbf{x}_i|\bar{\pi}(x_i)\big) = 0$.*

**Corollary 1** *If $\mathcal{Q}$ is definite then a run of Predict is complete.*

## 3.2 Stratified Completeness

The method that we propose in this section, called *Scomplete*, is based on "artificially blocking" the set of backpaths in a given network by assuming that the required nodes are believed. In essence the method is analogous to the algorithm of cutset-conditioning for updating probabilities in probabilistic Bayesian networks [23], except that it takes advantage of the soundness properties of *Predict* to proceed in stages. The procedure uses the structure of the network to find a set of nodes $\mathcal{CS} = \{c_1, \ldots, c_m\}$ such that $\mathcal{CS}$ blocks all backpaths in the network. Then, for any given node $x$, $\mathbf{x}$ is in $PlSet(\mathbf{x})$ (i.e., $\kappa(\mathbf{x}) = 0$) iff there exists an instantiation of the nodes in $\mathcal{CS}$ such that $\kappa(\mathbf{c_1}, \ldots, \mathbf{c_n}) = 0$ and $\kappa(\mathbf{x}|\mathbf{c_1}, \ldots, \mathbf{c_n}) = 0$. This is so since $\kappa(\mathbf{x}) = \min_{\mathcal{CS}} \kappa(\mathbf{x}|\mathbf{c_1}, \ldots, \mathbf{c_n}) + \kappa(\mathbf{c_1}, \ldots, \mathbf{c_n})$.

In order to proceed in stages *Scomplete* starts with an initial set of nodes $\{c_1, \ldots, c_n\}$, that blocks a *subset* of all backpaths. Then, for each instance of these nodes, *Scomplete* calls *Predict* to compute a new $PlSet(x_i)$. In the next stage *Scomplete* increases the set of initial blocking nodes by adding $\{c_{n+1}, \ldots, c_m\}$, computes again a new $PlSet(x_i)$, and so on. Note that at each stage the result of this computation is always sound and is at least as complete as the previous stage. Eventually *Scomplete* finds a set of nodes blocking of all backpaths in the graph and the result of the computation will be a complete set of plausible values for each node in the network.

*Scomplete* is described in Figure 2. Steps 2 and 3 are essentially the procedure suggested in [27] for isolating loops in a network. If $\Gamma'$ is empty in Step 3, then there are no loops, and the computation of $PlSet(x_i)$ is complete. Otherwise, all the nodes that remain in the modified graph $\Gamma'$ are part of loops.[7] The new root nodes of the modified graph constitute a good set of blocking nodes for two reasons. First they are readily identifiable; second, they are irrelevant to each other given the set BSet (which contains believed nodes, as well as blocking nodes used in previous stages). Thus,

---

[6]It can be extended in various ways including *conditional irrelevance*. In Section 5 we briefly discuss the potential of this definition in terms of plausibility and belief and the main difference with a definition of independence based on either probabilistic or OMP notions (see also [13]).

[7]Moreover only those nodes that have more than one incoming arc and their descendants are source of incompleteness. The reason is that only those nodes have parents that were assumed to be irrelevant to one another and were not.

PROCEDURE *Scomplete*
**Input:** A network $\Gamma$, and a set of believed nodes BSet.
**Output:** A complete PlSet($x_i$) for each $x_i$ in $\Gamma$.

1. Let $\mathcal{CS}$ be an empty set of nodes, and let PlSet($x_i$) be empty for every node in $\Gamma$.

2. Let $\Gamma'$ be a copy of $\Gamma$; modify $\Gamma'$ by removing nodes in BSet and any outgoing arcs from nodes in BSet.

3. While no longer removal is possible, find nodes in $\Gamma'$ with single neighbor (only one incoming or outgoing arc); remove the nodes and arcs.

4. If $\Gamma'$ is empty STOP; else let $\mathcal{R}$ be the set of root nodes in $\Gamma'$ and let $\mathcal{CS} = \mathcal{CS} \cup \mathcal{R}$.

5. For each instantiation of $\mathcal{CS} = \{c_1, \ldots, c_n\}$ such that $\kappa(c_1, \ldots, c_n) = 0$ compute PlSet'($x_i$) using *Predict*; let PlSet($x_i$) = PlSet($x_i$) $\cup$ PlSet'($x_i$).

6. Let BSet' be the set of nodes such that PlSet($x_i$) contains one element; then let BSet = BSet $\cup$ BSet' $\cup \mathcal{CS}$; goto 2.

END *Scomplete*

Figure 2: *Computing a complete set of plausibility values for each node $x_i$ in an OMP network.*

let $\vec{X}$ be any instantiation of BSet, and let $\{r_1, \ldots, r_m\}$ be the root nodes in $\Gamma'$. $\kappa(\mathbf{r_1}, \ldots, \mathbf{r_m}|\vec{X}) = 0$ whenever $\kappa(r_i|\vec{X}) = 0$ for every $r_i$, $1 \leq i \leq m$.

Step 4 simply augments the set of blocking nodes, and Step 5 invokes *Predict* in order to compute the PlSet($x_i$) of each node. There is some small amount of bookkeeping involved as *Predict* must be invoked for each instantiation of the blocking nodes, and the set PlSet($x_i$) must be assembled from the plausible sets in each iteration. Step 6 simply augments BSet for the next loop-isolation iteration in Steps 2 and 3.

## 4   Approximate Probabilistic Inference

It is well known that, in general, probabilistic inference in Bayesian networks is NP-hard [4], and as our networks scale to meet the requirements of new and more challenging applications this inference will become intractable. This fact emphasizes the importance of having a combination of anytime capabilities [2] and approximation techniques. In this section we explore the use of *Predict* in two anytime probabilistic algorithms: bounded-conditioning proposed by Horvitz, Suermondt, and Cooper [19], and the search-based algorithm proposed by Poole [24]. Both these algorithms can improve their efficiency by using a fast procedure capable of estimating which events in the sampling space will have significant probabilities, and which events can be ignored in the computation. The idea is to take advantage of the connection between OMPs and probabilities through the parameter $\epsilon$, and use *Predict* to estimate, in polynomial time, whether $P(\mathbf{x}) > \epsilon$.

As mentioned in Section 2, the properties of OMPs and the properties of probability distributions are closely related when $\epsilon$ is infinitesimally removed from zero. Yet, when used in conjunction with probabilistic algorithms to estimate whether $P(x)$ is above a given threshold represented by $\epsilon$, this $\epsilon$ must take a real value. We introduce a formal definition of an $\epsilon$-OMP, one that results from the transformation of a numerical probability to a OMP based on a given $\epsilon$, where $\epsilon$ is a real value. We then study the consequences of computing with an $\epsilon$-OMP as if $\epsilon$ where an infinitesimal quantity.

**Definition 7** ($\epsilon$-OMP.) *Let* P *be a probability distribution represented by a network $\Gamma$ with nodes $\{x_1, \ldots, x_n\}$, and let $\epsilon$ be a real quantity bigger than zero. The $\epsilon$-OMP of* P *is defined as the OMP $\kappa(x_n, \ldots, x_1) = \sum_{1 \leq i \leq n} \kappa(x_i|\pi(x_i))$,[8] that results from the following transformation: $\kappa(x_i|\pi(x_i)) = K$, where $K$ is an integer such that $\epsilon^{K+1} < P(x_i|\pi(x_i)) \leq \epsilon^K$.*

Using Def. 7 we can, given a real valued $\epsilon$, abstract a probability distribution P to a particular $\epsilon$-OMP and use *Predict* to compute the plausible set of values for each variable.

When $\epsilon$ is an infinitesimal quantity, we can interpret the output of *Predict* as establishing that if $\mathbf{x}$ is in PlSet($\mathbf{x}$), then $P(\mathbf{x}) > \epsilon$. Yet, when $\epsilon$ is a real quantity, and as $\epsilon$ is bigger and farther removed from zero, we should expect an error with regards to the connection to probabilistic values. This error appears because infinitesimal quantities do not accumulate but real quantities do. Thus, lower $\epsilon$-terms (i.e., those with an exponent bigger than $K$) may be significant for computations in the probabilistic domain, even though they are ignored in the OMP domain. As an example consider a network $\Gamma$ of nodes $\{x_1, \ldots, x_n\}$ in the form of a "chain" where $x_n$ is the only root node, and the only parent of node $x_i$ is node $x_{i-1}$ for $2 \leq i \leq n$. Consider a quantification where $P(x_1) = 1 - \epsilon$, $P(x_i|x_{i-1}) = 1 - \epsilon$, and $P(x_i|\neg x_{i-1}) = \epsilon$, for some real valued $\epsilon$. Then it is easy to verify that $P(x_n) > (1 - \epsilon)^n > 1 - n\epsilon$. Using $\epsilon$, the $\epsilon$-OMP transformation of this distribution yields $\kappa(\neg x_1) = 1$, $\kappa(x_1) = 0$, and $\kappa(x_i|x_{i-1}) = \kappa(\neg x_i|\neg x_{i-1}) = 0$, $\kappa(\neg x_i|x_{i-1}) = \kappa(x_i|\neg x_{i-1}) = 1$ for $2 \leq i \leq n$, and $\kappa(x_n) = 0$. If the value of $\epsilon$ is such that $\epsilon > 1 - n\epsilon$, $\kappa(x_n)$ will still be equal to zero (namely plausible), yet $P(x_n) \not> \epsilon$. Note that this situation can also appear in a "bushy" network where node $y$ is a functional "AND" of its parents $\{x_1, \ldots, x_n\}$ with $P(x_i) = 1 - \epsilon$ for $1 \leq i \leq n$, since $P(y) > 1 - n\epsilon$, given that

$$P(y) = \sum_{x_1, \ldots, x_n} P(y|\mathbf{x_1}, \ldots, \mathbf{x_n}) \prod_{1 \leq i \leq n} P(x_i).^9$$

Thus, both the length of the largest path in a given network and the maximum number of parents in a given family influence the error between probabilities and OMPs. Further research is needed in order to provide a precise characterization of this error in terms of the structural properties of the network.

---

[8] Such that for at least one instantiation of the nodes in $\Gamma$, $\kappa(\mathbf{x_1}, \ldots, \mathbf{x_n}) = 0$.

[9] Thanks to K. Fertig and D. Koller for this example.

Sections 4.1 and 4.2 describe how to use predict to approximate probabilistic values with both the bounded conditioning algorithm and Poole's search-based algorithm. Results and an application to plan evaluation [14] are provided in Section 4.3.

### 4.1    Predict and Bounded Conditioning

The method proposed by Horvitz, Suermondt and Cooper [19] is based on the cutset conditioning algorithm for evaluating probabilistic Bayesian networks with cycles [23]. In cutset conditioning, dependency loops in a Bayesian network are "broken" by a set of nodes $\{c_1, \ldots, c_n\}$ called cutset $\mathcal{CS}$. An instantiation $c_1, \ldots, c_n$ of $\mathcal{CS}$ is called an instance of the cutset. The probability of a given node $x$ is determined by the following equation

$$P(x) = \sum_{c_1, \ldots, c_n} P(x|c_1, \ldots, c_n) P(c_1, \ldots, c_n). \quad (3)$$

The value of $P(x|c_1, \ldots, c_n)$ for each instance is normally computed using a very efficient algorithm such as the polytree algorithm [23]. The term $P(c_1, \ldots, c_n)$ can also be computed using the polytree algorithm (see [27] for details). Note that the complexity of the computation is determined by the number of instances of the cutset, times the complexity of the polytree algorithm. The number of instances of the cutset is determined by product of the number of values of each node $c_i$ in $\mathcal{CS}$. Clearly this number grows exponentially as we add nodes to $\mathcal{CS}$.

The idea behind bounded-conditioning is that in order to establish whether the probability of interest is above certain value it may not be necessary to complete the computation of Eq. 3. Each instance of $\mathcal{CS}$ can be considered as a separate subproblem, and the computation can be done subproblem by subproblem and then stopped as soon as the desired threshold is met. Let us use $w_i$ to denote an instance $c_1, \ldots, c_n$ of the cutset, where $1 \leq i \leq k$, and let $j$ denote the number of instances of Eq. 3 computed so far. Then the value of $P(x)$ is bounded as shown in the following equation [19]: $\sum_{1 \leq i \leq j} P(x|w_i)P(w_i) \leq P(x) \leq \sum_{1 \leq i \leq j} P(x|w_i)P(w_i) + \sum_{j+1 \leq i \leq k} P(w_i)$. Note that as more subproblems are evaluated the closer we get to the real value of $P(x)$, hence the anytime character of the method.

The role of Predict is in providing an estimate of the values of $P(w_i)$, so that those instances with bigger probability values are evaluated first. This is of course specially significant in real time applications where computations must respond to time constraints. The problem with computing this ranking of the instances using conventional methods is, of course, that this computation will amount to updating the probabilities of the network which is the problem in the first place.

Given an $\epsilon$ a run of Predict will return, in polynomial time, those values of $c_i$, where $c_i$ is a node in the cutset $\mathcal{CS}$, for which $P(c_i) < \epsilon$.[10] Since $P(c_i) > P(c_1, \ldots, c_i, \ldots, c_n)$ we know that we can eliminate from our initial computation of Eq. 3 all those instances of $\mathcal{CS}$ where $c_i$ is participant.

The benefits in using Predict as described above will depend on the probability distribution over the set $\mathcal{CS}$. A worst case will be when $P(w_i) = \frac{1}{m}$ for $1 \leq i \leq m$, since depending on $\epsilon$ Predict will either eliminate all instances of $P(w_i)$ or will include all these instances. A best case is when $P(w_i)$ is ordered in stratas of $\epsilon, \epsilon^2, \epsilon^3, \ldots$, since by using different values of $\epsilon$, Predict will be able to rank these instances efficiently and allow the computation of $P(x)$ using only a subset of the subproblems. Some experimental results are provided in Section 4.3.

### 4.2    Predict and Poole's Search-Based Algorithm

Given a Bayesian network, this algorithm builds a search tree in which each leaf of the the tree represents a possible instantiation of all the variables of the probability distribution. The problem of belief update in the network is transformed into a search problem in the expansion of this tree. The role of Predict is in providing a lookahead estimator of which branches of the tree can be pruned (according to the threshold imposed by the parameter $\epsilon$).

Consider a Bayes network with nodes $\{x_1, \ldots, x_n\}$, where the indexes of these nodes are ordered consistently with any topological sort of these nodes according to $\Gamma$. The root of the search tree is labeled with $<>$ and the children of a node labeled with $< x_1, \ldots, x_j >$ are the nodes labeled with $< x_1, \ldots, x_j, x_{j+1} >$ for each instantiation $x_{j+1}$ of node $x_{j+1}$. The leaves of the tree are tuples of the form $< x_1, \ldots, x_n >$. The probability of each node in the tree are given by the equation $P(x_1, \ldots, x_j) = \prod_{1 \leq i \leq j} P(x_i|\vec{\pi}(x_i))$.

The algorithm to generate this tree proceeds as follows. At any point we have a queue $Q$ with tuples representing partial instantiations $< x_1, \ldots, x_j >$. When a partial instantiation is selected, it is expanded with a new value $x_j$. If $i = n$ then we reached a leaf and the tuple is removed from the queue. Let us use $\omega$ to denote a complete instantiation of the variables $\{x_1, \ldots, x_n\}$. Suppose we want to compute $P(g)$. At any stage during the algorithm we can divide the set of complete instantiations $\Omega$ as those that have been already generated $W$, and those that will be gener-ated from the queue. Let $P_W^g = \sum_{\omega \in W \wedge \omega \models g} P(w)$ and $P(g) = \sum_{\omega \in \Omega \wedge \omega \models g} P(w)$. Then at anytime during the algorithm $P(g)$ is bounded by $P_w^g \leq P(g) \leq P_Q$ where $P_Q$ is the sum of the probability of the elements in the queue. Note that as the algorithm progresses $P_Q$ is a monotonically nondecreasing quantity and consequently the algorithm can be claimed to be anytime.

---

[10]Subject to the approximation discussed in the previous section.

The complexity of the algorithm is of course in direct proportion with the number of worlds that need to be enumerated in order to get a good approximation of $P(g)$. The role of *Predict* is precisely that of finding which worlds can be ignored in this computation. There are at least two strategies for using *Predict* to explore and prune the search space. The first strategy involves providing an $\epsilon$ and then running *Predict* on the network to provide all those instances $< \mathbf{x_1}, \ldots, \mathbf{x_n} >$ such that $P(\mathbf{x_i}) < \epsilon$ for $1 \leq i \leq n$. Note that care has to be taken not to eliminate all worlds relevant to $P(g)$, therefore it has to be the case that $P(g) > \epsilon$. The other strategy uses *Predict* to select which of the current instances in the queue can be ignored. For this strategy, given an $\epsilon$, we run *Predict* with a particular partial instantiation $< \mathbf{x_1}, \ldots, \mathbf{x_j} >$ to decide whether $P(g|\mathbf{x_1}, \ldots, \mathbf{x_j}) > \epsilon$. Note that $P(g, \mathbf{x_1}, \ldots, \mathbf{x_j}) = P(g|\mathbf{x_1}, \ldots, \mathbf{x_j}) * P(\mathbf{x_1}, \ldots, \mathbf{x_j})$, and as soon as either of these terms is less than $\epsilon$, the instance can be ignored for all practical purposes. Both these strategies rely on the fact that *Predict* provides a "lookahead" estimate of the uncertainty in polynomial time.

### 4.3   Experimental Results

We have explored the use of *Predict* as part of a probabilistic algorithm in the context of an application to plan evaluation [14]. In this application the user represents an uncertain domain (and its dynamics) using Bayesian networks and then is interested in evaluating and ranking the performance of different plans. Plans can be sequences of actions, conditional actions, or policies as in Markov decision processes (MDPs) [10]. The user may also use the same techniques to monitor and forecast the progress of a given plan.

Most of the time the user is interested in a lower bound of the probability of the variables of interest rather than in their exact value, specially if this approximation can be computed fast. In the context of this application the user provides a value for $\epsilon$ and then can use *Predict* and a version of the ideas discussed in Section 4.1 combined with other strategies due to Darwiche [7], to obtain different estimates about the degree of belief of variables of interest in the domain (we are currently extending the capabilities of the tool to include the computation of utility values). Note that if $\epsilon$ is small the values computed will be closer to the real value of $P(x)$. Yet the time required in reaching this value will be longer since more subproblems (in terms of Eq. 3) must be considered. The bigger the $\epsilon$, the faster the computation will go; yet, the poorer the approximation to $P(x)$.

These networks are particularly difficult for conventional probabilistic algorithms because they contain a large number of loops by virtue of the representation of time [11, 8]. The characteristics of the networks we experimented with in this application were as follows: The number of nodes was around the hundreds, the number of nodes required for a cutset was around twenty, and the number of states, that is the number of subproblems that need to be considered for an exact computation, was around $20 \times 10^6$. The big number of subproblems is due to the fact that some of the variables had big state spaces. The quantification of these networks contained extreme probabilities (.99), as well as more moderate ones (including .7, and .5).

The way we computed the error in the update due to the approximations, was to average the probabilistic mass lost for each variable. Since we were missing some subproblems due to the pruning done by *Predict*, the probabilities of each variable in the domain did not add up to one. The difference represented the loss of mass, a factor we denote by $LM$. Thus if the algorithm returned $P'(\mathbf{x}) = p$, then $p \leq P(\mathbf{x}) \leq p + LM$. Some of the results were: for $\epsilon = 0.2$, the algorithm computed an answer in approximately 2 minutes. Yet, the quality of the answer was not very good, on average, $LM = .5$. For $\epsilon = 0.1$ the computation took nearly 4 minutes, and the accuracy of the answers increased considerably, with $LM < .05$. For, $\epsilon = 0.01$, the computation took approximately 6 minutes, with $LM < .002$. The exact computation, considering all the subproblems, using the dynamic conditioning algorithm described in [7] took 9 hours. We remark that we do not intend to make any claims on the performance of exact algorithms from these timings. The objective is to show the pruning power of *Predict*, and illustrate the relation between $\epsilon$ and $LM$ for a particular case. Careful comparisons to other exact and approximate probabilistic algorithms such as clustering-based methods and stochastic simulation [22, 5] is the subject of future research (see Section. 5). The computations by *Predict* took approximately 50 milliseconds in all cases. The experiments were conducted on a Sparc-10, and all the algorithms were implemented in Common Lisp.

We performed further experiments with different networks to assess how many subproblems of the cutset-conditioning we have to evaluate in order to get a reasonable estimate of the probabilities of each node. This is important since it provides a direct metric of how useful and effective is the estimation that *Predict* computes. The networks considered where a set of plan-evaluation networks such as the ones described above, the well known alarm network (see [19]), and the car network described in [9, 18]. We modified the car network and expanded it over time, and in addition we considered evidence for faulty conditions in order to flatten the distribution over the cutset cases.

For the plan-evaluation networks and the car network we found that with less than 5% of the total subproblems the error in the probabilistic estimate will be on the third most significant digit (i.e. on average the $LM \approx .001$). For the plan-evaluation networks this required the evaluation of 4000 subproblems, and in the car network it required 32 subproblems. The total number of subproblems that need to be evaluated in the car network for an exact computation was approximately $36 \times 10^3$. The complete set of data shows that in both these cases the distribution over the subproblems is very favorable to the use of *Predict*. The alarm network in turn required approximately 60% of the subproblems (72 subproblems in absolute numbers). Yet, the total number of subproblems for the alarm network was 108, and an exact computation was feasible in less than one minute.

## 5   Discussion

In this paper we have mainly focused on formally introducing *Predict*, proving its main properties, and describing its interaction with probabilistic inference. Yet, in addition to providing fast estimates of probabilistic values, there are at least two other topics related to *Predict* that deserve further attention. The first is as providing a clear link between belief networks, conditional logics and formalisms for nonmonotonic reasoning, and the second is as introducing a concept of irrelevance in which both the structure of the network and its quantification play a role. We briefly examine these in turn, summarize the main results and describe future work.

**Belief update, conditional logics, and networks.** Even though *Predict* takes as input an OMP represented by a ranking $\kappa$, it currently only distinguishes between $\kappa(x) = 0$ and $\kappa(x) > 0$, which correspond to $\text{Pl}(x)$ and $\neg \text{Pl}(x)$. Indeed the whole algorithm and its output can be described in terms of two modalities: belief and plausibility [13]. This fact has two important consequences: first it establishes a link between belief networks and conditional logics of belief studied by Boutilier [3] and others in the context of belief revision and nonmonotonic reasoning, and second, it provides an algorithm for prediction the consequences of actions in these formalisms. From Def. 1 we have the following properties of plausibility and belief: (1) If $\neg \text{Pl}(\neg x|y)$ then $\text{Bel}(x|y)$ and, (2) $\text{Pl}(x \wedge y)$ iff $\text{Pl}(x|y)$ and $\text{Pl}(y)$. Using these constraints embedded in the structure of the network can be expressed in terms of plausibility as: $\text{Pl}(x_1, \ldots, x_n)$ iff $\text{Pl}(x_n|\pi(x_n))$ and... and $\text{Pl}(x_1|\pi(x_1))$). Additional work is required in order to find compact and natural ways to specify these constraints in a logical language, and to characterize the set of logical models that satisfy a theory expressed using networks of belief.

**Belief-based irrelevance and independence.** As we mentioned in Section 3.1, Def. 4 is preliminary. We can expand it by introducing the notion of a context and by expressing it in terms of plausibility and belief: We say that $x$ is irrelevant to $y$ in the context of $z$ whenever $\text{Pl}(x \wedge y)$ iff $\text{Pl}(x)$ and $\text{Pl}(y)$ whenever $\text{Bel}(z)$. Note the difference between this definition and similar definitions for irrelevance and independence in terms of probabilities or OMPs, where context is established by conditioning. In probabilities, for example $x$ is independent of $y$ given $z$ whenever $P(x, y|z) = P(x|z) * P(y|z)$ ($\kappa(x, y|z) = \kappa(x|z) + \kappa(y|z)$). The formal properties of a belief-based notion of irrelevance are yet to be determined. The potential of exploring this definition is twofold: first, it will be instrumental, in conjunction with a network-based representation, in formalizing a notion of irrelevance for logics of belief in nonmonotonic reasoning, and second, it could lead to extending *Predict* to deal with diagnostic reasoning, where evidence may be influential in the complexity of the computations.

**More on *Predict* and probabilistic inference.** Further experiments are necessary in order to characterize the efficiency and power of an heuristic estimator such as *Predict* in the context of cutset-conditioning related algorithms. In addition, a comparison to stochastic simulation methods [5] would provide good benchmarks in terms of speed of computation. Similar to *Predict* these methods are anytime, and very effective for tasks of prediction. Note however that in contrast to stochastic simulation methods where errors can only be estimated through randomized approximations, the computations returned by *Predict* are guaranteed to be lower bounds of the probability function. We remark that for these comparisons to be meaningful, care must be taken to isolate factors related to particular implementations and platforms. Finally, we are also investigating the combination of *Predict* with other algorithms for probabilistic inference including clustering-based methods [22],[11] D'Ambrosio's SPI algorithm [6], and other search methods such as the one proposed by Henrion [17].

**OMPs and probabilities.** The main idea behind *Predict* is based on ignoring dependencies in the computation of belief. This strategy opens an intriguing question: to what extend can we ignore undirected cycles in a Bayesian network, and use the values computed as heuristic estimates of the real probabilistic values? The analysis in this paper provides an answer in the context of prediction tasks in terms of OMPs. Yet, what is the meaning, if there is one, of ignoring loops in the general case?[12] A related problem is the study and investigation of the consequences of assuming a real valued $\epsilon$ as opposed to an infinitesimal quantity for an OMP. The examples presented in

---

[11] A first approach would be to follow the directions suggested in [21].

[12] B. D'Ambrosio has started experimenting with this technique in decision making problems related to diagnosis and repair with surprising results (personal communication).

Section 4 only scratch the surface. They indicate that this approximation degrades in terms of the structural properties of the network. More work is needed in order to provide a precise characterization.

**Summary.** With the exception of [16], previous work on OMPs, $\epsilon$-semantics, and $\kappa$-rankings have focused mainly on the representational benefits of these formalisms [15, 9, 18]. This paper shows the clear benefits of using OMPs for computing in Bayesian networks. As an algorithm *Predict* takes advantage of both the structure of the network and its quantification. We have also shown how can the relation between OMPs and probabilities, through the parameter $\epsilon$, can be exploited to approximate and speed up probabilistic inferences. The applications of *Predict* that we are currently exploring include: plan evaluation, plan monitoring, and forecasting for contingency planning in stochastic domains.

**Acknowledgments.** Many thanks to P. Dagum, T. Dean, K. Fertig, H. Geffner, and M. Peot for useful discussions on related topics, and to A. Darwiche, M. Druzdzel, N. Friedman and especially to D. Koller for comments on previous versions of this paper. This work was funded in part by ARPA contract # F30602-91-C-0031.

# References

[1] S. Benferhat, D. Dubois, and H. Prade. Representing default rules in possibilistic logic. In *Proc. of Principles of Knowledge Representation and Reasoning Conf. (KR-92)*, pages 673–684, 1992.

[2] M. Boddy and T. Dean. Decision-theoretic deliberation scheduling for problem solving in time-constrained environments. *Artificial Intelligence*, 67(2):245–286, 1994.

[3] C. Boutilier. Conditional logics of normality: A modal approach. *Artificial Intelligence*, 68:87–154, 1994.

[4] G. Cooper. The computational complexity of probabilistic inference using Bayesian belief networks. *Artificial Intelligence*, 42:393–405, 1990. Research note.

[5] P. Dagum and E. Horvitz. Bayesian analysis of simulation algorithms. *Networks*, 23:499–516, 1993.

[6] B. D'Ambrosio. Local expression languages for probabilistic dependence. In *Proc. of the 7th Conf. on Uncertainty in AI*, pages 95–102, 1991.

[7] A. Darwiche. Conditioning algorithms for exact and approximate inference in causal networks. In *Proc. of the 11th Conf. on Uncertainty in AI*, 1995. This volume.

[8] A. Darwiche and M. Goldszmidt. Action networks: A framework for reasoning about actions and change under uncertainty. In *Proc. of the 10th Conf. on Uncertainty in AI*, pages 136–144, 1994.

[9] A. Darwiche and M. Goldszmidt. On the relation between kappa calculus and probabilistic reasoning. In *Proc. of the 10th Conf. on Uncertainty in AI*, pages 145–153, 1994.

[10] T. Dean, L. P. Kaebling, J. Kirman, and A. Nicholson. Planning with deadlines in stochastic domains. In *Proc. of the American Association for Artificial Intelligence Conf.*, pages 574–579, 1993.

[11] T. Dean and K. Kanazawa. A model for reasoning about persistence and causation. *Computational Intelligence*, 5(3):1442–150, 1989.

[12] H. A. Geffner. *Default Reasoning: Causal and Conditional Theories*. MIT Press, Cambridge, MA, 1992.

[13] M. Goldszmidt. Belief-based irrelevance and networks: Toward faster algorithms for prediction. Working notes: AAAI Fall Symposium on Relevance, 1994.

[14] M. Goldszmidt and A. Darwiche. Plan simulation using Bayesian networks. In *11th IEEE Conf. on Artificial Intelligence Applications*, pages 155–161, 1995.

[15] M. Goldszmidt and J. Pearl. Rank-based systems: A simple approach to belief revision, belief update, and reasoning about evidence and actions. In *Proc. of Principles of Knowledge Representation and Reasoning Conf. (KR-92)*, pages 661–672, 1992.

[16] M. Goldszmidt and J. Pearl. Reasoning with qualitative probabilities can be tractable. In *Proc. of the 8th Conf. on Uncertainty in AI*, pages 112–120, 1992.

[17] M. Henrion. Search-based methods to bound diagnostic probabilities in belief nets. In *Proc. of the 7th Conf. on Uncertainty in AI*, pages 142–150, 1991.

[18] M. Henrion, G. Provan, B. del Favero, and G. Sanders. An experimental comparison of diagnostic performance using infinitesimal and numerical Bayesian belief networks. In *Proc. of the 10th Conf. on Uncertainty in AI*, pages 319–326, 1994.

[19] E. Horvitz, G. Cooper, and H. Suerdmont. Bounded conditioning: Flexible inference for decisions under scarce resources. In *Proc. of the 5th Workshop on Uncertainty in AI*, pages 182–193, 1989.

[20] D. Hunter. Parallel belief revision. In Shachter, Levitt, Kanal, and Lemmer, editors, *Uncertainty in Artificial Intelligence (Vol. 4)*, pages 241–252. North-Holland, Amsterdam, 1990.

[21] F. Jensen and S. Andersen. Approximations in Bayesian belief universes for knowledge-based systems. In *Proc. of the 6th Conf. on Uncertainty in AI*, pages 162–169, 1990.

[22] F. Jensen, S. Lauritzen, and K. Olesen. Bayesian updating in causal probabilistic networks by local computations. *Computational Statistics Quarterly*, 4:269–282, 1990.

[23] J. Pearl. *Probabilistic Reasoning in Intelligent Systems: Networks of Plausible Inference*. Morgan Kaufmann, San Mateo, CA, 1988.

[24] D. Poole. The use of conflicts in searching Bayesian networks. In *Proc. of the 9th Conf. on Uncertainty in AI*, pages 359–367, 1993.

[25] P. Shenoy. On Spohn's rule for revision of beliefs. *International Journal of Approximate Reasoning*, 5(2):149–181, 1991.

[26] W. Spohn. Ordinal conditional functions: A dynamic theory of epistemic states. In W. L. Harper and B. Skyrms, editors, *Causation in Decision, Belief Change, and Statistics*, pages 105–134. Reidel, Dordrecht, Netherlands, 1988.

[27] H. Suermondt and G. Cooper. Probabilistic inference in multiply connected networks using loop cutsets. *International Journal of Approximate Reasoning*, 4:283–306, 1990.